\documentclass[10pt,twocolumn,letterpaper]{article}

\usepackage{iccv}
\usepackage{times}
\usepackage{epsfig}
\usepackage{graphicx}
\usepackage{amsmath}
\usepackage{amssymb}
\usepackage{color}
\usepackage{multicol}
\usepackage{multirow}
\usepackage{booktabs}
\usepackage{arydshln}
\usepackage{pifont}
\newcommand{\cmark}{\ding{51}}%
\newcommand{\xmark}{\text{\ding{55}}}

\usepackage[pagebackref=true,breaklinks=true,letterpaper=true,colorlinks,bookmarks=false]{hyperref}
\usepackage[normalem]{ulem}
\usepackage[capitalize]{cleveref}

\iccvfinalcopy 

\def\iccvPaperID{3726} 

\ificcvfinal\fi

\begin{document}

\title{VideoFlow: Exploiting Temporal Cues for Multi-frame Optical Flow Estimation}

\author{
Xiaoyu Shi$^{1,2}$ \and
Zhaoyang Huang$^{1,2*}$ \and
Weikang Bian$^{1}$ \and
Dasong Li$^{1}$ \and Manyuan Zhang$^{1}$ \and
Ka Chun Cheung$^{2}$ \and
Simon See$^{2}$ \and
Hongwei Qin$^{3}$ \and
Jifeng Dai$^{4}$ \and
Hongsheng Li$^{1,5,6}$\thanks{Corresponding author: Zhaoyang Huang and Hongsheng Li} \\ \and
$^{1}$Multimedia Laboratory, The Chinese University of Hong Kong \and
$^{2}$NVIDIA AI Technology Center \and
$^{3}$SenseTime Research \and
$^{4}$Tsinghua University \and
$^{5}$Centre for Perceptual and Interactive Intelligence (CPII) \and
$^{6}$Shanghai AI Laboratory \and
\{xiaoyushi@link, drinkingcoder@link, hsli@ee\}.cuhk.edu.hk
}


\maketitle
\ificcvfinal\thispagestyle{empty}\fi

\begin{abstract}
We introduce VideoFlow, a novel optical flow estimation framework for videos.
In contrast to previous methods that learn to estimate optical flow from two frames, VideoFlow concurrently estimates bi-directional optical flows for multiple frames that are available in videos by sufficiently exploiting temporal cues.

We first propose a TRi-frame Optical Flow (TROF) module that estimates bi-directional optical flows for the center frame in a three-frame manner.
The information of the frame triplet is iteratively fused onto the center frame.
To extend TROF for handling more frames, we further propose a MOtion Propagation (MOP) module that bridges multiple TROFs and propagates motion features between adjacent TROFs.
With the iterative flow estimation refinement, the information fused in individual TROFs can be propagated into the whole sequence via MOP.
By effectively exploiting video information, VideoFlow presents extraordinary performance, ranking 1st on all public benchmarks.
On the Sintel benchmark, VideoFlow achieves 1.649 and 0.991 average end-point-error (AEPE) on the final and clean passes, a 15.1\% and 7.6\% error reduction from the best published results (1.943 and 1.073 from FlowFormer++). On the KITTI-2015 benchmark, VideoFlow achieves an F1-all error of  3.65\%, a 19.2\% error reduction from the best published result (4.52\% from FlowFormer++). Code is released at \url{https://github.com/XiaoyuShi97/VideoFlow}.
\end{abstract}

\section{Introduction}



\begin{figure}[t]
    \centering
    \resizebox{1.0\linewidth}{!}{
\setlength{\tabcolsep}{0pt}
    \includegraphics[width=0.5\textwidth, trim={75mm 60mm 40mm 25mm}, clip]{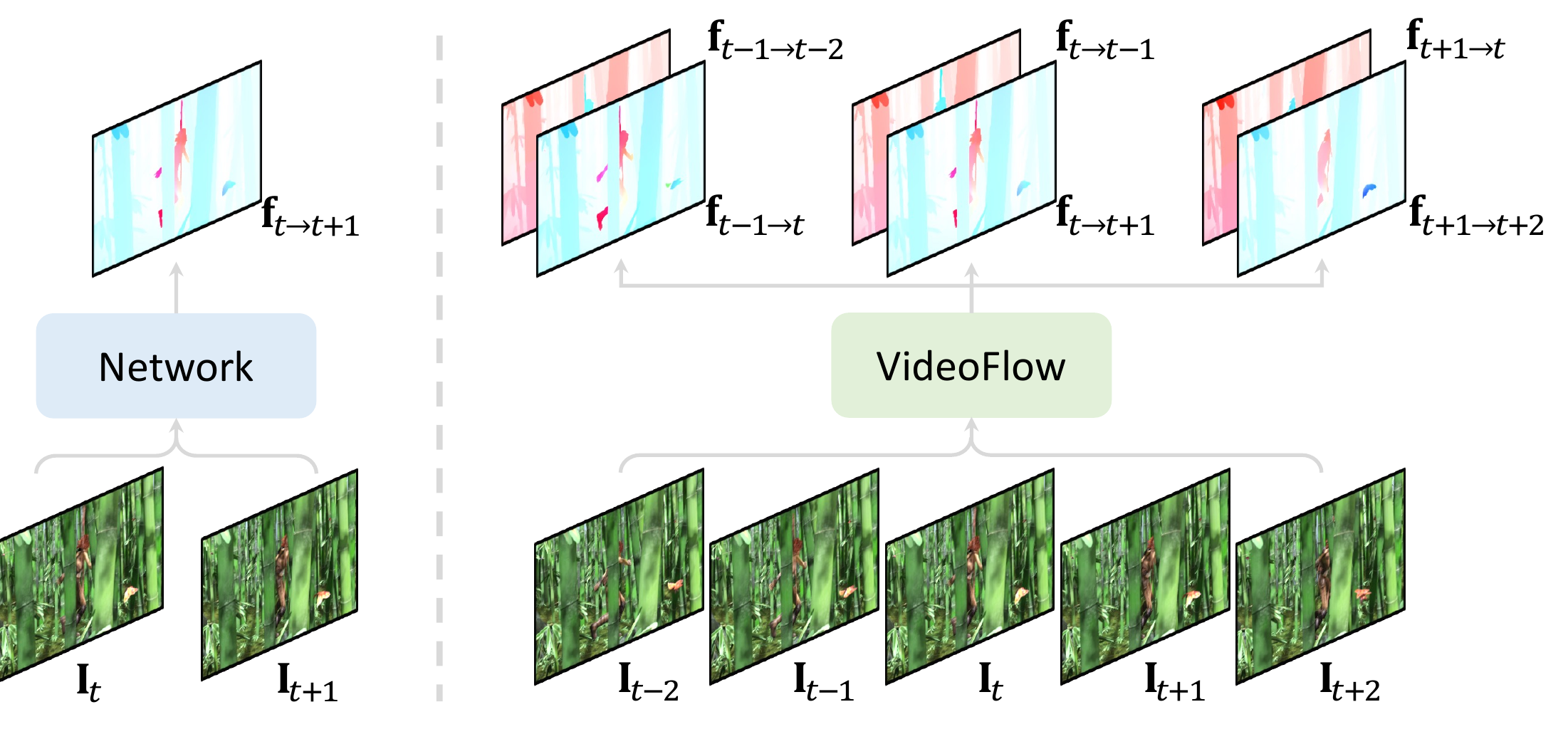}
    }
    \caption{\textbf{Comparison between two-frame and multi-frame optical flow estimation.} (Left) Previous methods are limited to optical flow estimation of frame pairs. (Right) We introduce VideoFlow, a novel framework that concurrently estimates bi-directional optical flows for multiple consecutive frames.}
    \label{fig:2f_flow}
\end{figure}

Optical flow estimation is a fundamental computer vision task of estimating pixel-wise displacement fields between consecutive video frames. It is widely adopted to tackle various downstream video problems, including video restoration~\cite{kim2019deep,xu2019deep,gao2020flow,lai2017deep,chan2021basicvsr,sajjadi2018frame,zhang2023kbnet,Li2022Learning}, video object detection~\cite{zhu2018towards,Wang_2018_ECCV,zhu2017flow}, video synthesis~\cite{yang2023rerender,hu2023dynamic,hu2023videocontrolnet,wu2023better,wu2023human} and action recognition~\cite{sun2018optical,piergiovanni2019representation,zhao2020improved}, providing valuable information on motion and cross-frame correspondence.

With the evolvement of dedicated datasets~\cite{ilg2017flownet} and model designs~\cite{huang2022flowformer,teed2020raft}, optical flow estimation has been greatly advanced over the years. 
However, we observe a divergence in its development from downstream demands. On the one hand, despite multiple frames being available in video streams, most efforts in this field are limited to flow estimation based on frame pairs, thereby ignoring valuable temporal cues from additional neighboring frames.
Notably, a simple strategy for temporal information fusion, \textit{e.g.}, the ``warm-start'' in RAFT~\cite{teed2020raft}, brings non-trivial performance gain.
On the other hand, multi-frame bi-directional optical flows are imperative in many downstream video processing algorithms~\cite{chan2021basicvsr,chan2022basicvsr++,niklaus2020softmax,xu2019deep,li2021neural,zhang2023unified,huang2020rife,Li2022efficient,Li_2023_CVPR,liu2021fuseformer,liu2021decoupled}.
However, due to the lack of appropriate multi-frame optical flow models, existing algorithms have to repeatedly estimate flows in a pair-wise manner.
This highlights the need for optical flow models specifically designed for multi-frame scenarios.

In this paper, we introduce VideoFlow, as shown in Fig.~\ref{fig:2f_flow}, a novel framework that concurrently estimates optical flows for multiple consecutive frames. VideoFlow consists of two novel modules: 1) a TRi-frame Optical Flow module (TROF) that jointly estimates bi-directional optical flows for three consecutive frames in videos, and 2) a MOtion Propagation (MOP) module that splices TROFs for multi-frame optical flow estimation.

Specifically, we treat three-frame optical flow estimation as the basic unit for the multi-frame framework. We argue that the  center frame of the triplet plays the key role of bridging temporal information, which motivates two critical designs of our proposed TROF model. Firstly, we propose to jointly estimate optical flows from the center frame to its two adjacent previous and next frames, which ensures the two flows originate from the same pixel and belong to the same continuous trajectory in the temporal dimension. 
Some previous three-frame methods~\cite{teed2020raft,ren2019fusion} rely on warping flow estimation from the preceding frame pair to facilitate the estimation of the current frame pair. 
The key difference arises in the presence of occlusion and out-of-boundary pixels, where the warped preceding predictions and current predictions might belong to entirely different objects. 
Such misalignment wastes valuable temporal cues and even introduces erroneous motion information. Secondly, TROF comprehensively integrates the bi-directional motion information in a recurrent updating process.

After constructing the strong three-frame model, we further propose a MOtion Propagation (MOP) module that extends our framework to handle more frames. This module passes bi-directional motion features along the predicted ``flow trajectories". Specifically, in the recurrent updating decoder, MOP warps the bi-directional motion features of each TROF unit to its adjacent units according to current predicted bi-directional optical flows. The temporal receptive field grows as the recurrent process iterates so that our VideoFlow gradually utilizes wider temporal contexts to optimize all optical flow predictions jointly.
This brings a significant advantage that the ambiguity and inadequate information in two-frame optical flow estimation will be primarily reduced when the information from multiple frames is sufficiently employed.
For example, estimating optical flows for regions that are occluded or out of view in target images is too challenging in the two-frame setting but can be effectively improved when we take advantage of additional information from more contextual frames.

In summary, our contributions are four-fold:
1) We propose a novel framework, VideoFlow, that learns to estimate optical flows of videos instead of image pairs.
2) We propose TROF, a simple and effective basic model for three-frame optical flow estimation.
3) We propose a dynamic MOtion Propagation (MOP) module that bridges TROFs for handling multi-frame optical flow estimation.
4) VideoFlow outperforms previous methods by large margins on all benchmarks.

\section{Related Work}

\noindent \textbf{Optical flow.}
Optical flow estimation traditionally is modeled as an optimization problem that maximizes the visual similarity between image pairs with regularization terms \cite{horn1981determining,black1993framework,bruhn2005lucas,sun2014quantitative}. 
FlowNet \cite{dosovitskiy2015flownet} is the first method that end-to-end learns to regress optical flows with a convolutional network.
FlowNet2.0 \cite{ilg2017flownet} takes this step further, adopting a stacked architecture with the warping operation, which performs on par with state-of-the-art (SOTA) optimization-based methods. 
The success of FlowNets motivates researchers to design better network architectures for optical flow learning. 
A series of works, represented by SpyNet \cite{ranjan2017optical}, PWC-Net \cite{sun2018pwc,sun2019models}, LiteFlowNet \cite{hui2018liteflownet,hui2020lightweight} and VCN \cite{yang2019volumetric}, emergencies, employing coarse-to-fine and iterative estimation methodology. 
Despite the fast progress, these models inherently suffer from missing small fast-motion objects in the coarse stage and can only refine flows in limited steps. To remedy this issue, Teed and Deng \cite{teed2020raft} propose RAFT~\cite{teed2020raft}, which estimates optical flow in a coarse-and-fine (i.e. multi-scale search window in each iteration) and recurrent manner.
The accuracy is significantly improved along with the recurrent iterations increasing.
The iterative refinement paradigm is adopted in the following works~\cite{jiang2021learning,xu2021high,jiang2021learning2,zhang2021separable,hofinger2020improving}.
Recently, transformer-based architectures~\cite{huang2022flowformer,shi2023flowformer++,huang2023flowformer,li2023blinkflow,xu2022gmflow}  for optical flow, such as FlowFormer~\cite{huang2022flowformer}, shows great superiority against previous CNNs.
FlowFormer++~\cite{shi2023flowformer++,huang2023flowformer} further unleashes the capacity of transformers by pertaining with the Masked Cost-Volume Autoencoding (MCVA).
Optical flow is also extended to more challenging settings, such as low-light~\cite{zheng2020optical}, foggy~\cite{yan2020optical}, and lighting variations~\cite{huang2022neuralmarker}.

\noindent \textbf{Multi-frame optical flow.}
In the traditional optimization-based optical flow era, researchers used Kalman filter~\cite{elad1998recursive,chin1994probabilistic} to estimate optical flows with the temporal dynamics of motion for multi-frame optical flow estimation.
``warm-start''~\cite{teed2020raft}, which simply warps the flows of the previous image pairs as the initialization for the next image pairs, improves RAFT series~\cite{sui2022craft,sun2022skflow} by non-trivial margins.
PWC-Fusion~\cite{ren2019fusion}, which fuses information from previous frames with a GRU-RCN at the bottleneck of U-Net, only achieves 0.65\% performance gain over PWC-Net due to the rough feature encoding.
ContinualFlow~\cite{neoral2018continual}, Unsupervised Flow~\cite{Janai2018ECCV} and Starflow~\cite{godet2021starflow} only warp flow/features in one direction. ProFlow~\cite{maurer2018proflow}, similar to~\cite{ren2019fusion}, focuses on the three-frame setting, individually predicting
bi-directional flows and combining them with fusion model.
~\cite{hur2021self} utilizes LSTM to propagate motion information, which shows inferior performance than gradually increasing receptive field during recurrence in our experiment. SelfFlow~\cite{liu2019selflow} takes three-frame estimation as strong pretraining for two-frame
estimation.
In contrast, we integrate motion features from both forward and backward directions during flow refinement. Point-tracking~\cite{doersch2022tapvid,harley2022particle,bian2023context,wang2023tracking} is also closely related to multi-frame optical flow. In contrast to optical flow measuring dense correspondences between adjacent frames, point-tracking cares about the trajectories of the specified points on the following frames.

\begin{figure*}[t]
    \centering
    \resizebox{1.0\linewidth}{!}{
    \includegraphics[width=0.5\textwidth, trim={10mm 40mm 0mm 35mm}, clip
    ]{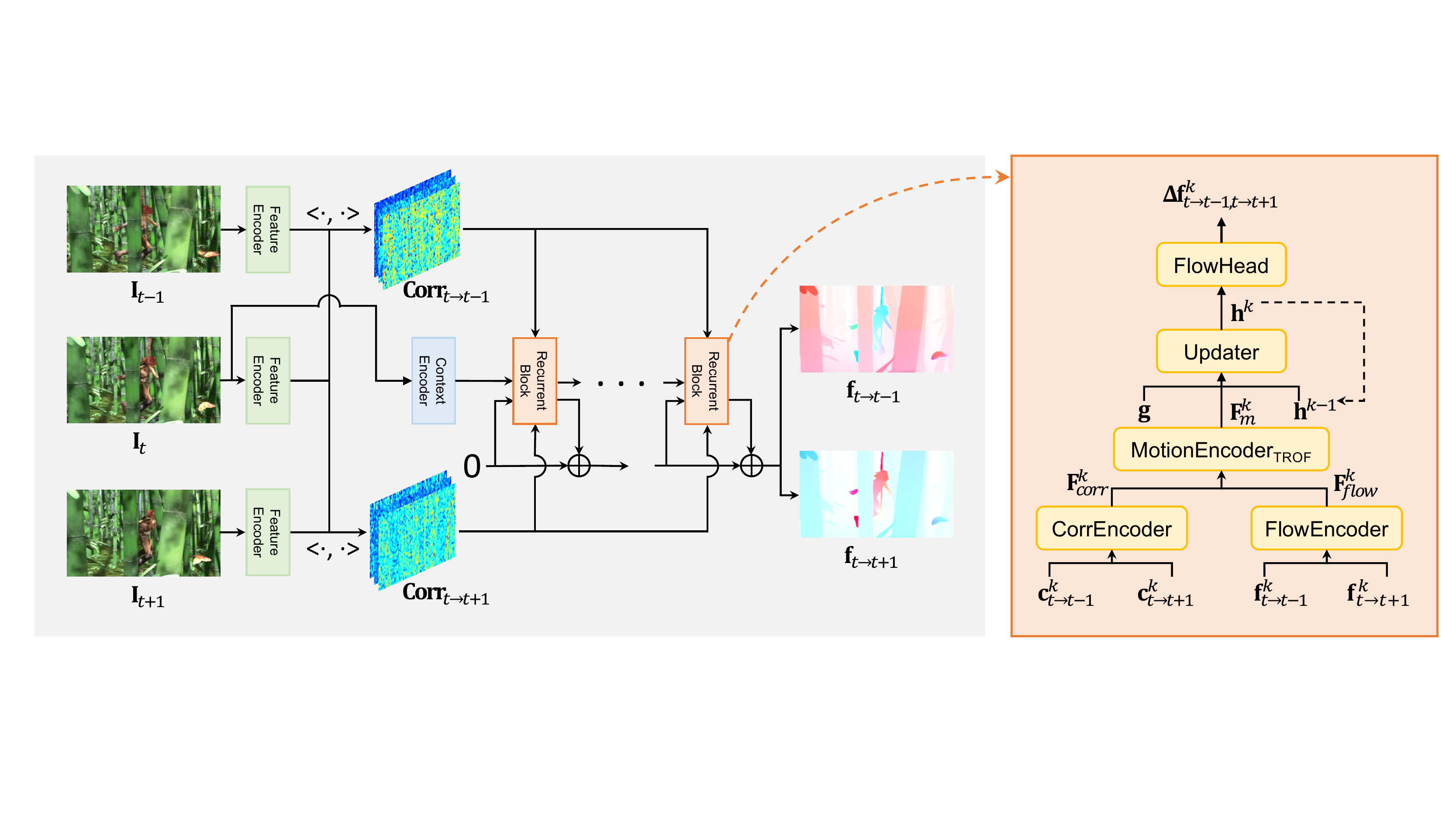}
    }
    \caption{\textbf{Overview of VideoFlow in the three-frame setting.} Given a triplet of frames as input, VideoFlow jointly estimates bi-directional optical flows from the center frame to the adjacent previous and next frames. After building dual cost volumes, it recurrently fuses bi-directional flow features and correlation features to update flow predictions. The orange block on the right illustrates the recurrent flow refinement block.}
    \label{fig:arch}
\end{figure*}

\section{Method}



Optical flow estimation targets at regressing a per-pixel displacement field $\mathbf{f}_{t \rightarrow t+1}: \mathbb{I}^{H\times W\times 2} \rightarrow \mathbb{R}^{H\times W\times 2}$, which maps each source pixel $\mathbf{x} \in \mathbb{I}^{2}$ of image $\mathbf{I}_t$ at time $t$ to its corresponding coordinate $\mathbf{x}'_{t+1} = \mathbf{x} + \mathbf{f}_{t \rightarrow t+1}(\mathbf{x})$ on the target image $\mathbf{I}_{t+1}$. Existing methods typically focus on the two-frame setting, thereby ignoring the rich temporal cues from more neighboring frames. In view of this limitation, we propose VideoFlow, a novel framework to exploit the valuable information from wider temporal context to achieve more accurate optical flow estimation.
VideoFlow mainly consists of two modules: 1) a TRi-frame Optical Flow (TROF) module designed to estimate bi-directional optical flows of three-frame clips,
and 2) a MOtion Propagation (MOP) module that extends TROF for multi-frame optical flow estimation.
In this section, we will elaborate the TROF and MOP modules.

\subsection{TROF for Tri-frame Optical Flow Estimation}
TROF learns to estimate bi-directional optical flows in a three-frame setting.
As shown in Fig.~\ref{fig:arch}, given a frame $\mathbf{I}_t$, its previous frame $\mathbf{I}_{t-1}$ and next frame $\mathbf{I}_{t+1}$, TROF iteratively estimates a sequence of bidirectional flows ${\mathbf f}^k \in \mathbb{R}^{H\times W\times 2\times 2}$, where $k=1,2,...,N$ indicates the refinement iteration step. $\mathbf{f}^{k}$ includes a 2D flow to the previous frame ${\mathbf{f}}^k_{t \rightarrow t-1}$ and a 2D flow to the next frame ${\mathbf{f}}^k_{t \rightarrow t+1}$. For brevity, we omit the subscript $t$ when the variables stands for the center frame $\mathbf I_t$.

\noindent \textbf{Dual Correlation Volumes from Tri-frame Features.}
Correlation volumes, which measure pixel-wise visual similarities between image pairs, provide pivotal information for flow estimation in previous methods~\cite{teed2020raft,huang2022flowformer,shi2023flowformer++}.
TROF also infers flows from the correlation volumes but it concurrently estimates bi-directional flows from the center frame to its previous frame $\mathbf{f}^k_{t \rightarrow t-1}$ and next frame $\mathbf{f}^k_{t \rightarrow t+1}$ by formulating a dual correlation volume.
Specifically, we encode three input images with a feature encoder that outputs a feature map of shape $H\times W\times D$ for each frame. $H$ and $W$ are the feature height and width at $1/8$ resolution of original images, and we set feature dimension $D=256$. TROF builds dual correlation volumes $\mathbf{Corr}_{t,t-1}, \mathbf{Corr}_{t,t+1}\in \mathbb R^{H\times W\times H\times W}$ by computing their pixel-wise dot-product similarities. Given the dual correlation volume, TROF concurrently predicts the bi-directional flows.

\noindent \textbf{Bi-directional Motion Feature Fusion.}
The core of our TROF lies in fully fusing bi-directional motion information to the center frame. 
Similar to RAFT~\cite{teed2020raft}, TROF iteratively retrieves multi-scale correlation values $\mathbf{Corr}(\mathbf f)$ centered at current flows to refine the bi-directional optical flow.
Specifically, at the $k$-th iteration, we retrieve correlation values $\mathbf c^k_{t\rightarrow t-1} = \mathbf{Corr}_{t,t-1}(\mathbf f^k_{t\rightarrow t-1})$ and $\mathbf c^k_{t\rightarrow t+1} = \mathbf{Corr}_{t,t+1}(\mathbf f^k_{t\rightarrow t+1})$ from dual correlation volumes according to currently predicted bi-directional flows ${\mathbf{f}}^{k}_{t \rightarrow t-1}$ and ${\mathbf{f}}^{k}_{t \rightarrow t+1}$ respectively. 
We fuse and encode correlation feature $\mathbf{F}^k_{corr} \in \mathbb{R}^{H\times W \times D_c}$, flow feature $\mathbf{F}^k_{flow} \in \mathbb{R}^{H\times W \times D_f}$ at the center frame as
\begin{equation}
\begin{aligned}
    & \mathbf{F}^k_{corr}={\rm CorrEncoder}(\mathbf c^k_{t\rightarrow t-1},\mathbf c^k_{t\rightarrow t+1}), \\
    & \mathbf{F}^k_{flow}={\rm FlowEncoder}(\mathbf{f}^{k}_{t\rightarrow t-1},\mathbf{f}^{k}_{t\rightarrow t+1}). \\
\end{aligned}
\label{Eq: correlation fusion}
\end{equation}
The correlation values $\mathbf c^k_{t\rightarrow t-1}, \mathbf c^k_{t \rightarrow t+1}$ from the backward and forward flow fields are passed to a correlation encoder to obtain fused correlation features $\mathbf{F}^k_{corr}$.
Similarly, we also fuse current predicted bi-directional flows ${\mathbf{f}}^{k}_{t \rightarrow t-1}$ and ${\mathbf{f}}^{k}_{t \rightarrow t+1}$ with the flow encoder to obtain the flow features $\mathbf{F}^k_{flow}$.
The correlation feature $\mathbf{F}^k_{corr}$ and flow feature $\mathbf{F}^k_{flow}$ respectively encode the dual correlation values and bi-directional displacement information.
Then we further encode $\mathbf{F}^k_{corr}$ and $\mathbf{F}^k_{flow}$ to obtain the bi-directional motion feature $\mathbf{F}^k_m \in \mathbb{R}^{H\times W \times D_m}$:
\begin{equation}
\begin{aligned}
    & \mathbf{F}^k_{m}={\rm MotionEncoder_{TROF}}(\mathbf F^k_{corr}, \mathbf F^k_{flow}).
\end{aligned}
\label{Eq: motion encoder}
\end{equation}
The information from the dual correlation volumes is well aligned at the center frame because the bi-directional flow fields originate from the same source pixels to the two neighboring frames.
$\mathbf{F}^k_m$ contains rich motion and correlation information and is used to predict residual bi-directional flows as introduced below.

\noindent \textbf{Bi-directional Recurrent Flow Updating.} 
Following RAFT~\cite{teed2020raft},
we maintain a hidden state $\mathbf{h}^k$ to cache features for recurrent flow refinement.
We also use the visual features $\mathbf g$ of the center frame $\mathbf I_t$ as the context feature and initialize the hidden state with the context feature as $\mathbf{h}^0=\mathbf g$.
By passing the motion feature $\mathbf{F}^k_m$, along with the context feature $\mathbf g$ and hidden state feature $\mathbf h^{k}$ from the previous iteration into a recurrent updating block to update the hidden state, we decode bidirectional residual flows $\mathbf{\Delta f}^k \in \mathbb{R}^{H\times W \times 2 \times 2}$ from the hidden state:
\begin{equation}
\begin{aligned}
    & \mathbf{h}^{k+1} = {\rm Updater}(\mathbf{F}^k_m, \mathbf g, \mathbf h^{k}), \\
    & \mathbf{\Delta f}^k_{t\rightarrow t-1, t\rightarrow t+1} = {\rm FlowHead}(\mathbf{h}^{k+1}), \\
    & {\mathbf{f}}^{k+1}_{t\rightarrow t-1, t\rightarrow t+1} = {\mathbf{f}}^k _{t\rightarrow t-1, t\rightarrow t+1}+ \Delta \mathbf{f}^k_{t\rightarrow t-1, t\rightarrow t+1} .\\
\end{aligned}
\label{Eq: raft-style update}
\end{equation}
The predicted flow residuals $\mathbf{\Delta f}^k$ is added to the currently predicted flow to iteratively refine the flow prediction.
For all the encoders, we use the SKBlock following SKFlow~\cite{sun2022skflow}, which consists of large-kernel depth-wise convolution layers (details in the supplemental material). 


\begin{figure*}[t]
    \centering
    \resizebox{1.0\linewidth}{!}{
    \includegraphics[width=0.5\textwidth, trim={42mm 20mm 42mm 50mm}, clip
    ]{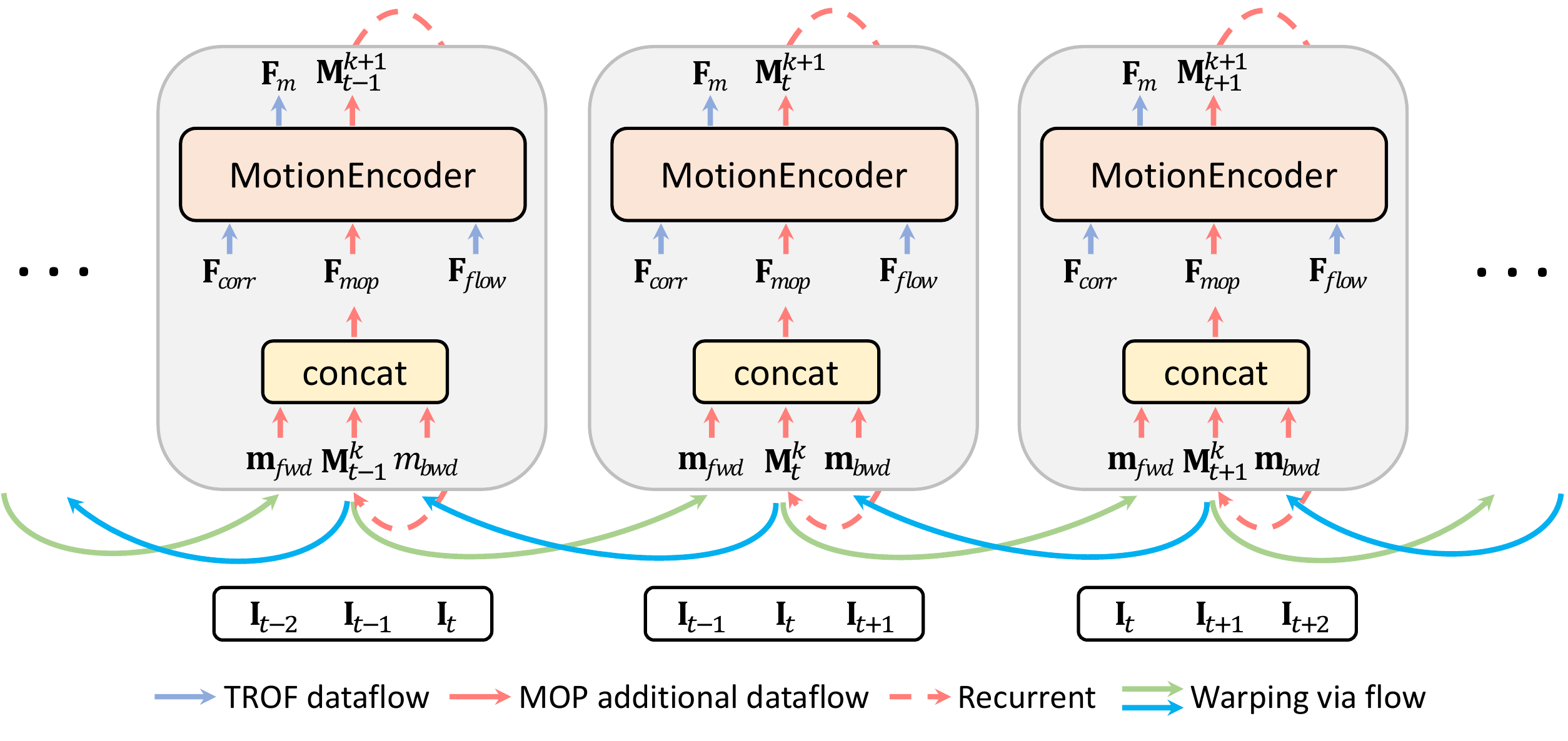}
    }
    \caption{\textbf{MOP motion encoder.} We bridge TROF units with the motion propagation module for tackling optical flow estimation of more than three frames. Specifically, we additionally maintains a motion state feature $\mathbf M_t^k$ in the motion encoder of each TROF unit. In each iteration, the motion state features of adjacent TROF units are warped to the center frame as auxiliary information. $\mathbf M_t^k$ is updated by the motion encoder and exploits wider temporal contexts as the recurrent process iterates.}
    \label{fig:TROF+MOP}
\end{figure*}

\subsection{Bridging TROFs with Motion Propagation}

TROF is a simple and effective module for three-frame bi-directional flows.
We further extend TROFs for tackling optical flow estimation of more than three frames with a MOtion Propagation (MOP) module.
Give multiple frames in a video as input, they are split into overlapped frame triplets and a TROF is in charge of handling a triplet.
In Fig.~\ref{fig:TROF+MOP}, we illustrate processing five input frames and use three TROFs on the triplets of $\{\mathbf{I}_{t-2},\mathbf{I}_{t-1},\mathbf{I}_{t}\}$, $\{\mathbf{I}_{t-1},\mathbf{I}_{t},\mathbf{I}_{t+1}\}$,$\{\mathbf{I}_{t},\mathbf{I}_{t+1},\mathbf{I}_{t+2}\}$.
However, such a design does not take advantage of the extra temporal context information as the temporal fusion is limited within each single TROF. 
To efficiently propagate information across neighboring TROFs, MOP improves the motion encoder in TROF (Eq.~(\ref{Eq: motion encoder})) by recurrently fusing motion features warped from adjacent TROFs.
Therefore, the motion features in each TROF are gradually propagated to the entire sequence via recurrent iterations.

Specifically, we additionally maintain a motion state feature $\mathbf M^k_t \in \mathbb{R}^{H\times W \times D_m}$ for the TROF unit on the triplet $\{\mathbf{I}_{t-1},\mathbf{I}_{t},\mathbf{I}_{t+1}\}$
in the motion encoder of MOP. 
$\mathbf M_t^0$ is randomly initialized and learned via training.
It serves to communicate motion features with adjacent TROFs centered at $t-1$ and $t+1$ and is updated to integrate temporal cues.
In each iteration $k$, we retrieve temporal motion information $\mathbf{m}^{k}_{fwd}$, $\mathbf{m}^{k}_{bwd}$ from adjacent TROFs by warping their motion state features $\mathbf M^k_{t-1}$ and $\mathbf M^k_{t+1}$ according to currently predicted flows:
\begin{equation}
\begin{aligned}
    & \mathbf{m}^{k}_{fwd} = \rm{Warp}(\mathbf{M}^k_{t+1};\mathbf{f}^{k}_{t \rightarrow t+1}), \\
    & \mathbf{m}^{k}_{bwd} = \rm{Warp}(\mathbf{M}_{t-1}^{k};\mathbf{f}^{k}_{t \rightarrow t-1}), \\
    & \mathbf{F}^k_{mop} ={\rm Concat}(\mathbf{M}_{t}^{k}, \mathbf{m}^{k}_{fwd}, \mathbf{m}^{k}_{bwd}). \\
\end{aligned}
\label{Eq: MOP update}
\end{equation}
After concatenating them with the motion state feature $\mathbf{M}_{t}^{k}$ of the current frame, we obtain the motion propagation feature $\mathbf F_{mop}^k$, which augments $\mathbf F^k_{corr}$ and $\mathbf F^k_{flow}$ by summarizing motion features of consecutive three TROFs in each iteration. The motion feature state $\mathbf{M}_{t}^{k}$ is also updated in each iteration. 
\begin{equation}
\begin{aligned}
    & \mathbf{F}_{m}^{k}, \mathbf{M}^{k+1}_t={\rm MotionEncoder_{MOP}}(\mathbf F^k_{corr}, \mathbf F^k_{flow}, \mathbf{F}^{k}_{mop}), 
\end{aligned}
\label{Eq: MOP motion encoder}
\end{equation}
where $\mathbf{F}_{m}^{k}$ is passed to the updater to predict residual flows $\mathbf{\Delta f}^k$ as in Eq~(\ref{Eq: raft-style update}), and $\mathbf{M}^{k+1}_t$ will be used in the next iteration $k+1$ of Eq~(\ref{Eq: MOP update}).
As MOP absorbs the hidden motion features from adjacent TROFs at each iteration, the temporal receptive field of the hidden motion features $\mathbf{M}_{t}^{k}$ gradually increases as iterations proceed.
We thus can process and integrate more TROFs jointly to expand the effective timespan with MOP.
In practice, our network takes $5$ frames as input and is trained to output $3$ center frames' bi-directional flows during training. 
During inference, the network is used to predict $T$ frames' flow at a time for input video clips of length $T+2$.



\subsection{Training Loss}

VideoFlow requires consecutive multiple frames with corresponding ground truth bi-directional optical flows $\mathbf f_{gt,t\rightarrow t-1}, \mathbf f_{gt,t\rightarrow t+1}$ for training.
We train VideoFlow to predict $T$ bidirectional flow
with $\ell 1$ loss for all predicted flows defined as:
\begin{equation}
\begin{aligned}
\mathcal{L} = \sum_{t=1}^{T} \sum_{k=0}^{N} & \gamma^{N-k}||\mathbf f_{gt,t\rightarrow t-1} - \mathbf f^k_{t\rightarrow t-1}||_1 \\ & + \gamma^{N-k}||\mathbf f_{gt,t\rightarrow t+1} - \mathbf f^k_{t\rightarrow t+1}||_1,
\end{aligned}
\label{Eq: training loss}
\end{equation}
where $N$ (set as 12 in our experiments) is the number of recurrent steps and $\gamma$ is set as 0.85 to add higher weights on later predictions following RAFT~\cite{teed2020raft}.

\section{Experiments}
We evaluate our VideoFlow on the Sintel~\cite{butler2012naturalistic} and the KITTI-2015~\cite{geiger2013vision} benchmarks. VideoFlow outperforms all previous methods on both benchmarks and reduces the average end-point-error (AEPE) to a subpixel level on the clean pass of Sintel for the first time.

\noindent \textbf{Experimental setup.} We adopt the average end-point-error (AEPE) and Fl-All(\%) as the evaluation metrics. The AEPE denotes mean flow error over all valid pixels. The Fl-All computes the percentage of pixels with flow error larger than 3 pixels and over 5\% of ground truth. The Sintel dataset consists of two passes rendered from the same model. The clean pass is easier with smooth shading and specular reflections, while the final pass enables more challenging rendering settings including atmospheric effects, motion blur and camera depth-of-field blur.

\noindent \textbf{Implementation Details.} Following the FlowFormer series~\cite{huang2022flowformer, shi2023flowformer++}, we use the first two stages of ImageNet-pretrained Twins-SVT~\cite{chu2021twins} as the image encoder and context encoder and fine-tune the parameters. We follow SKFlow~\cite{sun2022skflow} to replace ConvGRU with SKBlocks as an iterative flow refinement module. Since FlyingChairs~\cite{dosovitskiy2015flownet} only contains two-frame training pairs, we skip it and directly pretrain VideoFlow on the FlyingThings~\cite{mayer2016large} dataset. For our three-frame model, we pretrain it on the FlyingThings dataset for 300k iterations (denoted as `T'). Then we finetune it on the data combined from FlyingThings, Sintel, KITTI-2015 and HD1K~\cite{kondermann2016hci} (denoted as `T+S+K+H') for 120k iterations. This model is submitted to the Sintel online test benckmark for evaluation. We further finetune the model on the KITTI-2015 dataset for 50k iterations. For our five-frame model, we follow the same schedule with fewer training iterations: 125k iterations on `T', 40k iterations on `T+S+K+H' and 25k iterations on `K'. We choose AdamW optimizer and one-cycle learning rate scheduler. Batch size is set as 8 for all stages. The highest learning rate is set as $2.5\times 10^{-4}$ for FlyingThings and $1.25\times 10^{-4}$ on other training datasets. Please refer to supplementary for more details.

\begin{table*}[t]
\centering
\scriptsize
\resizebox{0.9\linewidth}{!}{
\begin{tabular}{clcccccccc}
\hline
\multicolumn{1}{c}{\multirow{2}{*}{Training Data}} & \multicolumn{1}{c}{\multirow{2}{*}{Method}} & \multicolumn{2}{c}{Sintel (train)}     & \multicolumn{2}{c}{KITTI-15 (train)}                    & \multicolumn{2}{c}{Sintel (test)}                     & \multicolumn{1}{c}{KITTI-15 (test)} \\
\cmidrule(r{1.0ex}){3-4}
\cmidrule(r{1.0ex}){5-6}
\cmidrule(r{1.0ex}){7-8}
\cmidrule(r{1.0ex}){9-9} 
\multicolumn{1}{c}{}                               & \multicolumn{1}{c}{}
& \multicolumn{1}{c}{Clean} & \multicolumn{1}{c}{Final} & \multicolumn{1}{c}{Fl-epe} & \multicolumn{1}{c}{Fl-all} & \multicolumn{1}{c}{Clean} & \multicolumn{1}{c}{Final} & \multicolumn{1}{c}{Fl-all} \\ 
\hline
\multirow{3}{*}{A}  &    Perceiver IO~\cite{jaegle2021perceiver} &  1.81 & 2.42 & 4.98  & - & - & - & - \\ 
&    PWC-Net~\cite{sun2018pwc}   & 2.17 & 2.91 & 5.76 & - & - & - & -  \\ 
& RAFT~\cite{teed2020raft}  &  1.95 & 2.57 &  4.23 & - & - & - & - \\
\hline
\multicolumn{1}{c}{\multirow{14}{*}{C+T}}           & HD3~\cite{yin2019hierarchical} & 3.84      & 8.77      & 13.17  & 24.0  & - & - & - \\
 & LiteFlowNet~\cite{hui2018liteflownet} & 2.48 & 4.04 & 10.39 & 28.5 & - & - & -  \\
 & PWC-Net~\cite{sun2018pwc} & 2.55 & 3.93 & 10.35 & 33.7 & - & - & - \\
 & LiteFlowNet2~\cite{hui2020lightweight} &  2.24 & 3.78 & 8.97 & 25.9 & - & - & - \\
 & S-Flow~\cite{zhang2021separable}  & 1.30 & 2.59 & 4.60 & 15.9 & & & \\
 & RAFT~\cite{teed2020raft}   & 1.43 & 2.71 & 5.04 & 17.4 & - & - & - \\
 & FM-RAFT~\cite{jiang2021learning2}   & 1.29 & 2.95 & 6.80 & 19.3 & - & - & -  \\
 & GMA~\cite{jiang2021learning}   &  1.30 & 2.74 & 4.69 & 17.1 & - & - & - \\
 & GMFlow~\cite{xu2022gmflow}   &  1.08 & 2.48 & - & - & - & - & - \\
 & GMFlowNet~\cite{zhao2022global}   & 1.14 & 2.71 & 4.24 & 15.4 & - & - & - \\
 & CRAFT~\cite{sui2022craft}   & 1.27 & 2.79 & 4.88 & 17.5 & - & - & - \\
 & SKFlow~\cite{sun2022skflow}   &1.22 & 2.46 & 4.47 & 15.5 & - & - & - \\
 & FlowFormer~\cite{huang2022flowformer} &$\uline{0.94}$ & 2.33 & 4.09 & 14.72 & - & - & - \\
 & FlowFormer++~$^\dag$\cite{shi2023flowformer++} & $\mathbf{0.90}$ & $\uline{2.30}$ & $\uline{3.93}$ & $\mathbf{14.13}$ & - & - & - \\
\hdashline
\multicolumn{1}{c}{\multirow{2}{*}{T}} & Ours (3 frames) & 1.03 & $\mathbf{2.19}$ & 3.96 & 15.33 & - & - & - \\
& Ours (5 frames) & 1.18 & 2.56 & $\mathbf{3.89}$ & $\uline{14.20}$ & - & - & - \\
\hline
\multirow{17}{*}{C+T+S+K+H}                         &    LiteFlowNet2~\cite{hui2020lightweight} & (1.30) & (1.62) & (1.47) & (4.8) & 3.48 & 4.69 & 7.74 \\
 &  PWC-Net+~\cite{sun2019models} &(1.71) & (2.34) &  (1.50) &  (5.3) & 3.45 &  4.60 & 7.72\\
 & VCN~\cite{yang2019volumetric} & (1.66) & (2.24) & (1.16) & (4.1) & 2.81 & 4.40 & 6.30 \\
 &  MaskFlowNet~\cite{zhao2020maskflownet} &  - & - & - & - & 2.52 & 4.17 & 6.10 \\
  &  S-Flow~\cite{zhang2021separable} & (0.69) & (1.10) & (0.69) & (1.60) & 1.50 & 2.67 & 4.64 \\
 &  RAFT~\cite{teed2020raft} & (0.76) & (1.22) & (0.63) & (1.5) & 1.94 & 3.18 & 5.10
\\
 &  FM-RAFT~\cite{jiang2021learning2} & (0.79) & (1.70) & (0.75) & (2.1) & 1.72 & 3.60 &  6.17 \\
 &  GMA~\cite{jiang2021learning} & - & - & - & - & 1.40 & 2.88 & 5.15 \\

  &  GMFlow~\cite{xu2022gmflow} & - & - & - & - & 1.74 & 2.90 & 9.32 \\
  &  GMFlowNet~\cite{zhao2022global} & (0.59) & (0.91) & (0.64) & (1.51) & 1.39 & 2.65 & 4.79 \\
  &  CRAFT~\cite{sui2022craft} & (0.60) & (1.06) & (0.57) & (1.20) & 1.45 & 2.42 & 4.79 \\
&  FlowFormer~\cite{huang2022flowformer} & (0.48) & (0.74) & (0.53) & (1.11) & 1.16 & 2.09 & 4.68 
\\
& FlowFormer++~$^\dag$\cite{shi2023flowformer++} & (0.40) & (0.60) & (0.57) & (1.16) & 1.07 & 1.94 & 4.52  \\
\cdashline{2-9}
&  PWC-Fusion*~\cite{sun2019models} &  - & - & - & - & 3.43 & 4.57 & 7.17\\
&  RAFT*~\cite{teed2020raft} & (0.77) & (1.27) & - & - & 1.61 & 2.86 & 5.10
\\
&  GMA*~\cite{jiang2021learning} & (0.62) & (1.06) & (0.57) & (1.2) & 1.39 & 2.47 & 5.15 \\
&  SKFlow*~\cite{sun2022skflow} & (0.52) & (0.78) & (0.51) & (0.94) & 1.28 & 2.23 & 4.84 \\
\hdashline
\multicolumn{1}{c}{\multirow{2}{*}{T+S+K+H}} & Ours (3 frames) & (0.37) & (0.54) & (0.52) & (0.85)  & $\uline{1.00}$ & $\uline{1.71}$ & $\uline{4.44}$ \\
& Ours (5 frames) & (0.46) & (0.66) & (0.56) & (1.05) & $\mathbf{0.99}$ & $\mathbf{1.65}$ & $\mathbf{3.65}$ \\
\hline \\
\end{tabular}
}
\caption{\label{Tab: comparison} $\textbf{Experiments on Sintel~\cite{butler2012naturalistic} and KITTI~\cite{geiger2013vision} datasets.}$ `A' denotes the autoflow dataset. `C + T' denotes training only on the FlyingChairs and FlyingThings datasets. `+ S + K + H' denotes finetuning on the combination of Sintel, KITTI, and HD1K training sets.  * denotes methods that use three frames for prediction. PWC-Fusion~\cite{ren2019fusion} fuses two independently predicted flows. Other methods use the warm-start strategy~\cite{teed2020raft}, which warps the estimation of the preceding frame pair to initialize the current estimation. $^\dag$ denotes that FlowFormer++~\cite{shi2023flowformer++} has an additional pre-training stage. We use $\mathbf{bold}$ and $\uline{~~~}$ to highlight the methods that rank 1st and 2nd.
}
\end{table*}

\subsection{Quantitative Experiment}
As shown in Table~\ref{Tab: comparison}, we evaluate VideoFlow on the Sintel~\cite{butler2012naturalistic} and KITTI-2015~\cite{geiger2013vision} benchmarks. Specifically, we compare the generalization performance of models on the training set of Sintel and KITTI-2015 (denoted as `C+T' for other two-frame models and `T' for our VideoFlow). We then compare the dataset-specific fitting ability of optical flow models after dataset-specific finetuning (denoted as `C+T+S+K+H' for other two-frame models and `T+S+K+H' for our VideoFlow). `A' refers to pre-training models on another synthetic dataset Autoflow~\cite{sun2021autoflow}, while its training code is not publicly available.

\noindent \textbf{Generalization Performance.} In Table~\ref{Tab: comparison}, the `T'/`C+T' settings reflect the cross-dataset generalization ability of models. Our VideoFlow achieves comparable performance with FlowFormer series~\cite{huang2022flowformer, shi2023flowformer++} and outperforms other models. It is worth noting that FlowFormer and FlowFormer++ have 35$\%$ more parameters than VideoFlow (18.2M vs 13.5M). FlowFormer++ is additionally pretrained with masked autoencoding strategy. Specifically, 3-frame VideoFlow ranks first on the challenging final pass of Sintel training set. The 5-frame VideoFlow achieves best performance on KITTI-2015 Fl-epe metric and is only second to FlowFormer++ on the Fl-all metric.

\noindent \textbf{Dataset-specific Performance.} After training our VideoFlow in the `T+S+K+H' setting, we submit it to the online Sintel benchmark. As shown in Table~\ref{Tab: comparison}, our 3-frame model already outperforms all published methods, achieving 1.00 and 1.71 AEPE on the clean and final passes, respectively. 5-frame VideoFlow further increases the accuracy. Specifically, it achieves 0.99 and 1.65 AEPE on the clean and final passes, a 7.6\% and 15.1\% error reduction from FlowFormer++, with much fewer parameters.
Then we further finetune VideoFlow on the KITTI-2015 training set and submit it to the online benchmark. The 3-frame VideoFlow achieves an Fl-all error of 4.44\%, surpassing all previous published methods. Our five-frame VideoFlow further obtains 3.65\%, a 19.2\% error reduction from the previous best-published method FlowFormer++.

\begin{figure*}
\centering
    \resizebox{1.0\linewidth}{!}{
\setlength{\tabcolsep}{2pt}
\begin{tabular}{@{} c c c @{}}
        \includegraphics[width=.32\linewidth, trim={0mm 40mm 0mm 0mm}, clip]{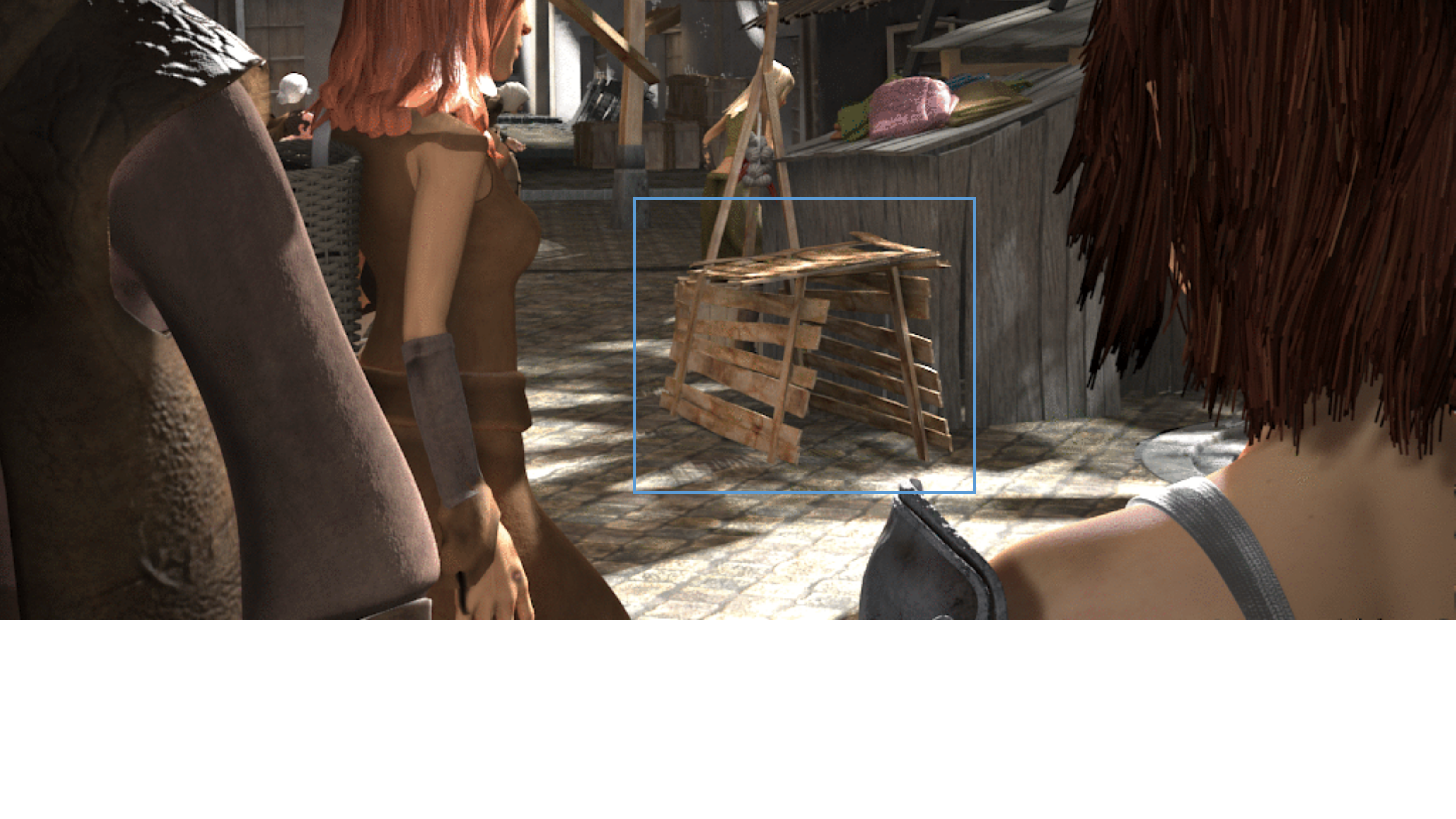} &
        \includegraphics[width=.32\linewidth, trim={0mm 40mm 0mm 0mm}, clip]{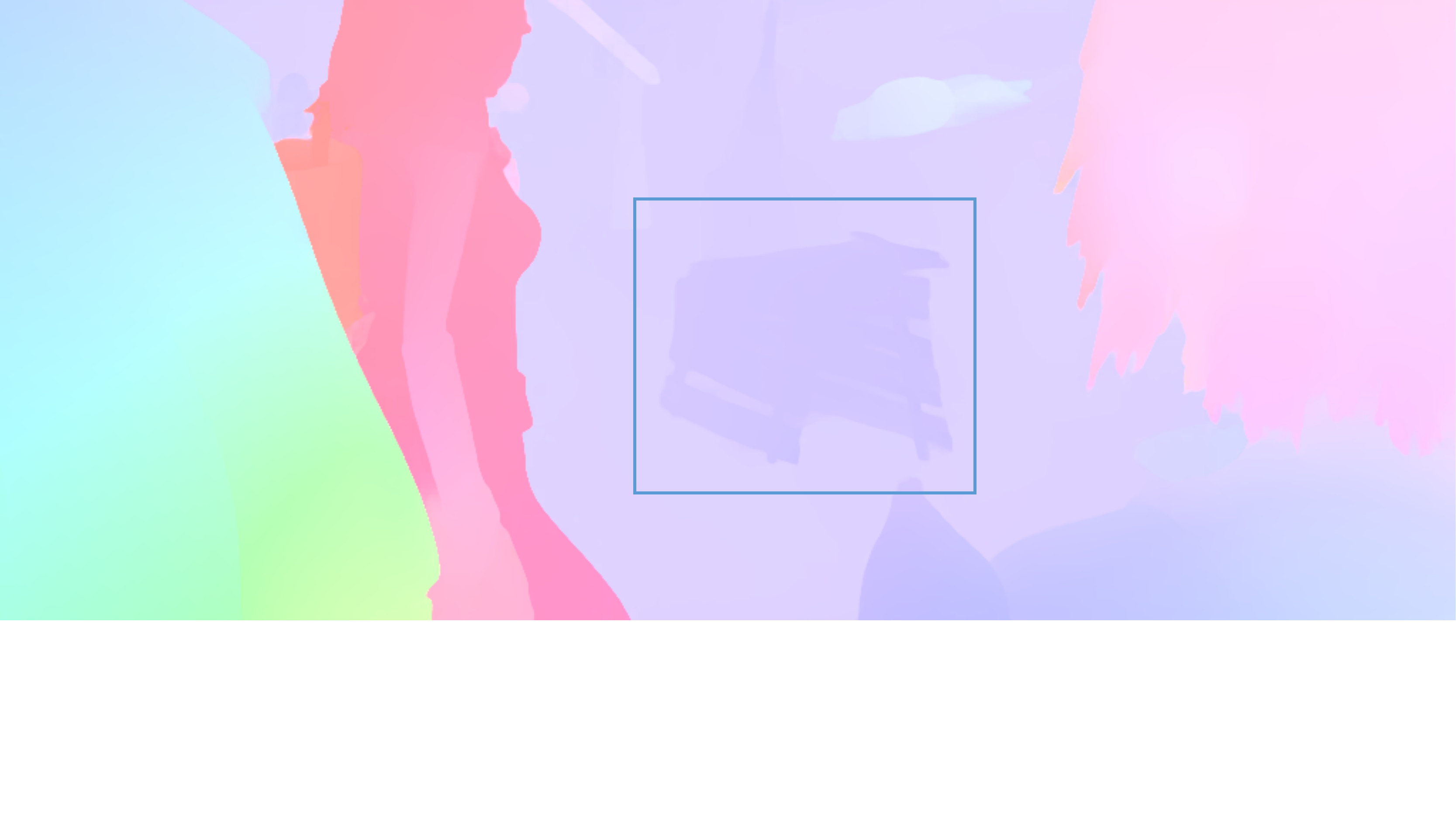} &
        \includegraphics[width=.32\linewidth, trim={0mm 40mm 0mm 0mm}, clip]{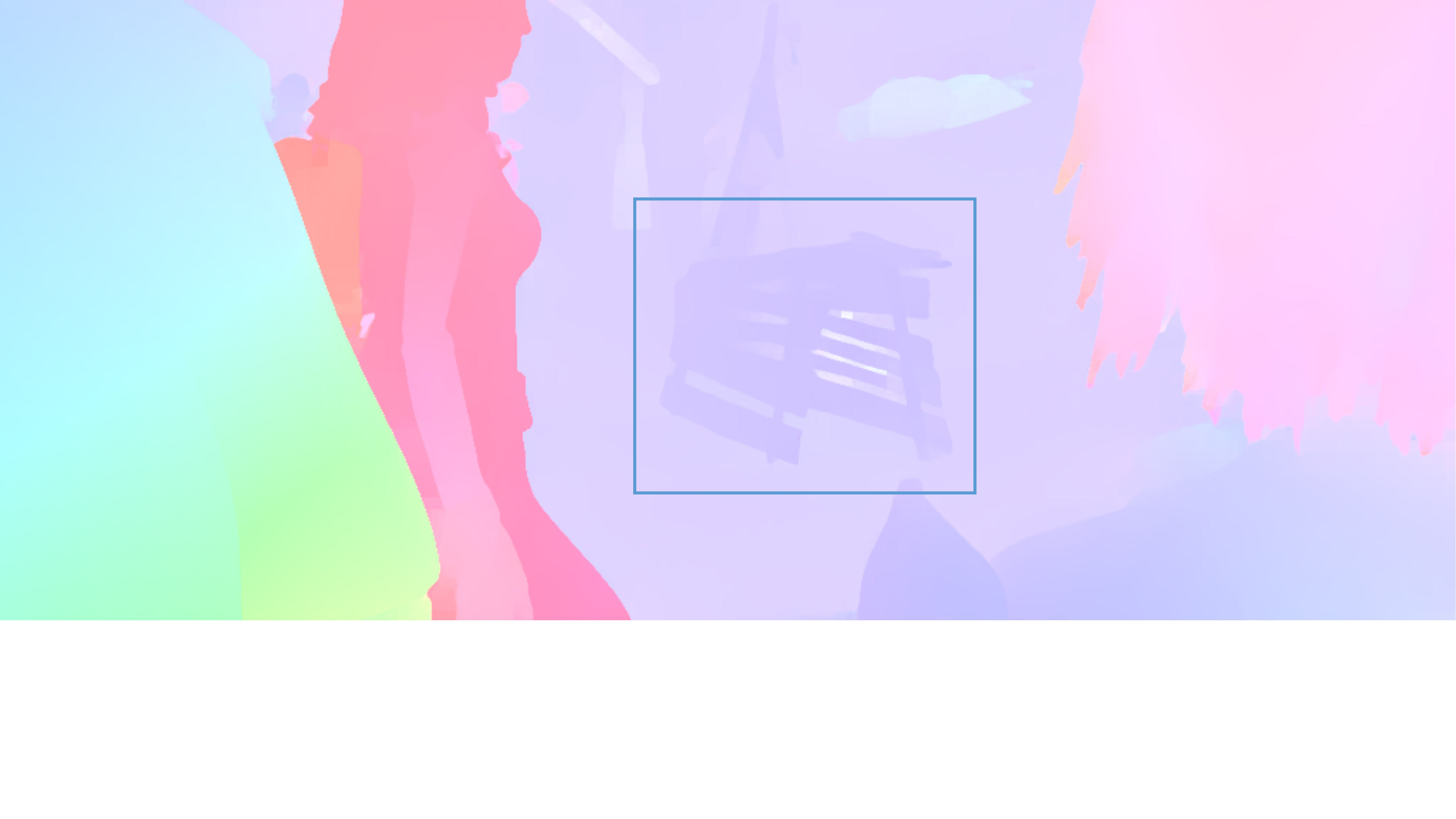}  \\
        \includegraphics[width=.32\linewidth, trim={0mm 80mm 0mm 0mm}, clip]{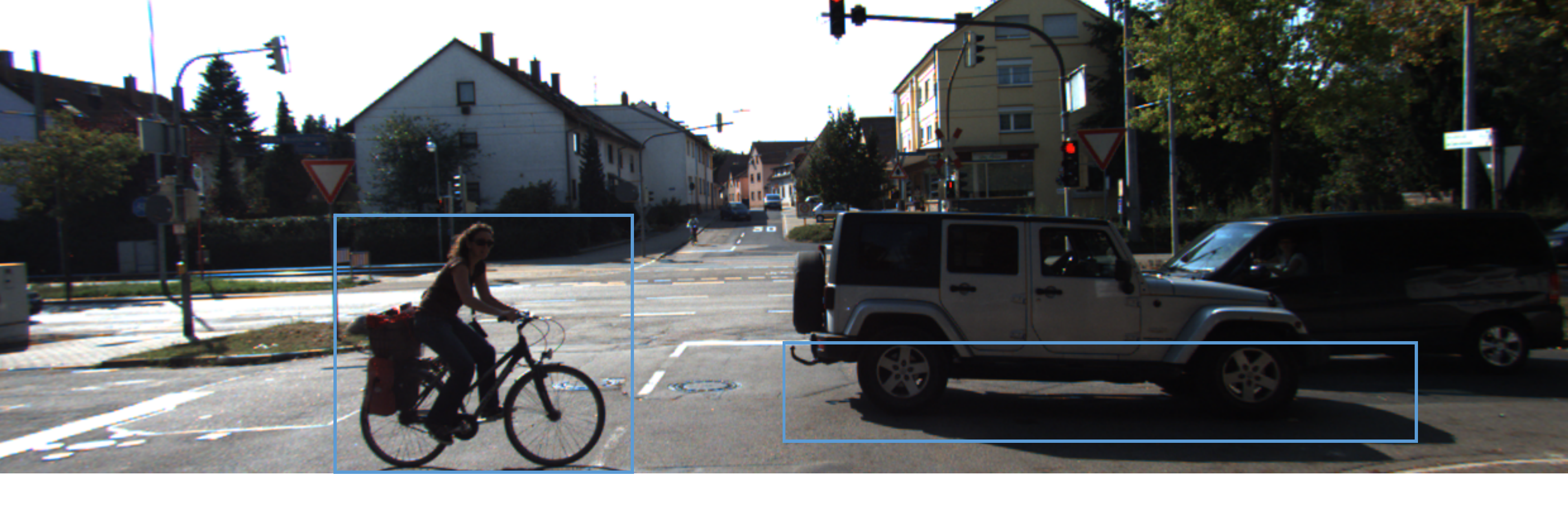} &
        \includegraphics[width=.32\linewidth, trim={0mm 80mm 0mm 0mm}, clip]{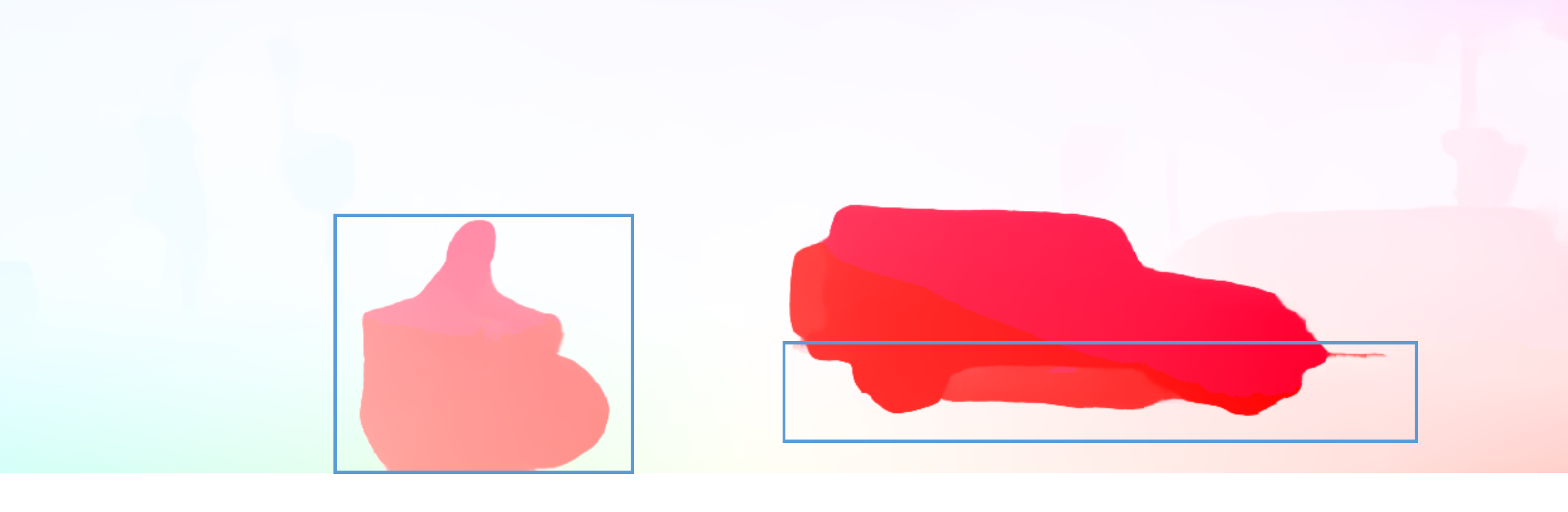} &
        \includegraphics[width=.32\linewidth, trim={0mm 80mm 0mm 0mm}, clip]{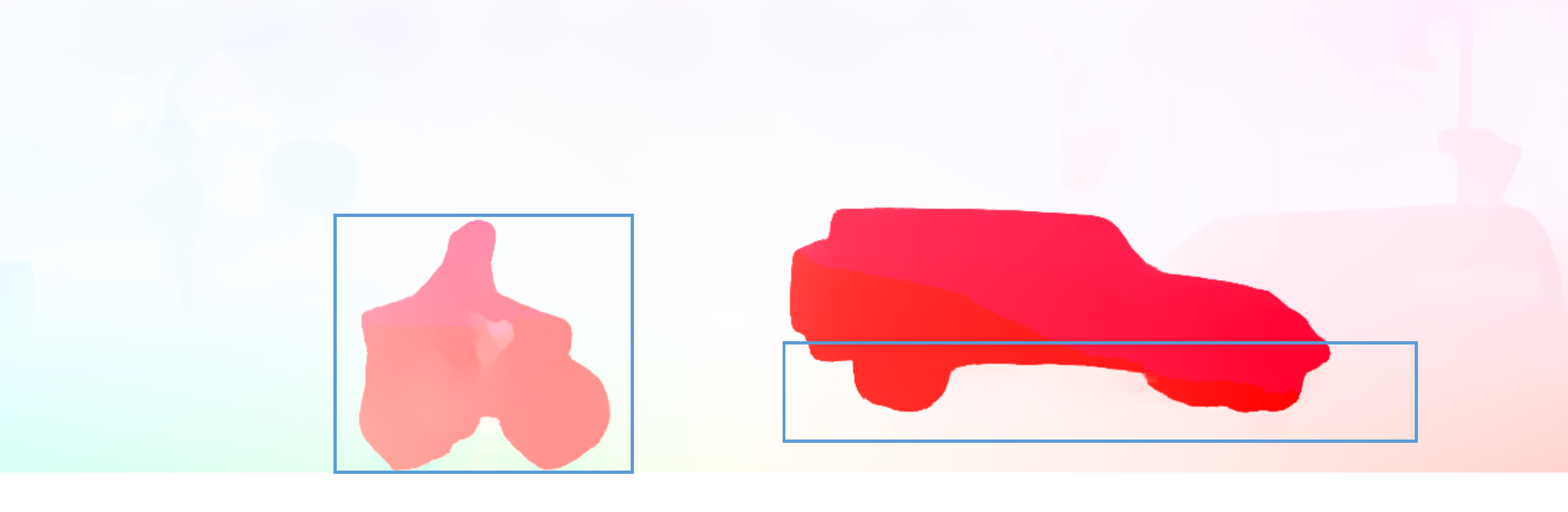} \\
        Input & FlowFormer++ & VideoFlow (Ours)  \\
    \end{tabular}
    }
    \caption{$\textbf{Qualitative comparison on Sintel and KITTI test sets.}$ VideoFlow preserves clearer details (row 1st).
    By better utilizing temporal cues, VideoFlow successfully distinguishes the ground from shadows and avoids accident artifacts (row 2nd).
    }
    \label{fig:quality}
\end{figure*}

\begin{table*}[t]
\centering
\setlength{\tabcolsep}{1pt}
\begin{tabular}{lcccccccccccc}

\hline
 \multicolumn{1}{c}{\multirow{2}{*}{Method}} & \multicolumn{6}{c}{Sintel Test (clean)}                    & \multicolumn{6}{c}{Sintel Test (final)} \\
\cmidrule(r{1.0ex}){2-7} 
\cmidrule(r{1.0ex}){8-13}
        & \multicolumn{1}{c}{All} & \multicolumn{1}{c}{Matched} & \multicolumn{1}{c}{Unmatched} & \multicolumn{1}{c}{$s_{0-10}$} & \multicolumn{1}{c}{$s_{10-40}$}& \multicolumn{1}{c}{$s_{40+}$}
        &\multicolumn{1}{c}{All} & \multicolumn{1}{c}{Matched} & \multicolumn{1}{c}{Unmatched} & \multicolumn{1}{c}{$s_{0-10}$} & \multicolumn{1}{c}{$s_{10-40}$}& \multicolumn{1}{c}{$s_{40+}$}\\
        
\hline
SKFlow*~\cite{sun2022skflow} & 1.28 & 0.57 & 7.25 & 0.28& 0.95& 7.17& 2.26 & 1.14 & 11.42 & 0.58 & 1.68 & 12.02 \\
FlowFormer~\cite{huang2022flowformer} & 1.16 & 0.42 & 7.16 & 0.26 & 0.82 & 6.44 & 2.09 & 0.96 & 11.30 & 0.46 & 1.47 & 11.66 \\
FlowFormer++~\cite{shi2023flowformer++} & 1.07 & $\uline{0.39}$ & 6.64 & 0.25 & $\uline{0.80}$ & 5.81 & 1.94 & 0.88 & 10.63 & 0.44 & 1.40 & 10.71 \\ 
VideoFlow (3 frames)  & $\uline {1.02}$ & $\mathbf {0.38}$ & $\uline {6.19}$ & $\mathbf{0.22}$ & $\mathbf{0.69}$ & $\uline{5.75}$ &$\uline{1.84}$ & $\uline{0.86}$ & $\uline{9.81}$ & $\uline{0.42}$ & $\uline{1.29}$ & $\uline{10.19}$ \\ 
VideoFlow (5 frames)  & $\mathbf {0.99}$ & 0.40 & $\mathbf {5.83}$ & $\uline{0.23}$ & $\mathbf{0.69}$ & $\mathbf{5.48}$ & $\mathbf{1.65}$ & $\mathbf{0.79}$ & $\mathbf{8.66}$ & $\mathbf{0.40}$ & $\mathbf{1.24}$ & $\mathbf{8.80}$ \\ 
\hline \\
\end{tabular}
\caption{\label{Tab: sintel detail}$\textbf{Sintel test results analysis.}$ 
 `Unmatched' refers to occluded or out-of-boundary pixels and $s_{0-10}$, $s_{10-40}$, $s_{40+}$ denote pixels with ground truth flow motion magnitude falling in $0-10$, $10-40$, and more than $40$ pixels, respectively. 
VideoFlow obtains clear improvements in challenging cases, including `Unmatched' pixels and pixels with large motions.
}
\end{table*}



\noindent \textbf{Multi-frame Methods Comparison.}
Besides VideoFlow, there are four other methods that can be regarded as estimating optical flow from three-frame information, i.e. PWC-Fusion~\cite{sun2019models}, RAFT*~\cite{teed2020raft}, GMA*~\cite{jiang2021learning}, and SKFlow*~\cite{sun2022skflow}.
PWC-Fusion adopts a temporal GRU at the bottleneck of PWC-Net~\cite{sun2018pwc} to fuse motion information from previous frames.
In the three-frame structure, the other three methods adopt the warm-start technique, which warps the flows from the former pair to the later pair as the initialization.
Such designs do not take information from future frames and only fuse information once at a coarse level, thereby bringing little benefits.
In contrast, VideoFlow deeply integrates information from both directions during iterative flow refinement.
3-frame VideoFlow outperforms PWC-Fusion by 70.3\% and SKFlow* by 20.3\% on clean pass.
Moreover, our 5-frame version further reduces the error (10.3\% on Sintel final pass and 21.6\% on KITTI), which beyond the capability of the warm-start technique because it can not draw benefits from longer sequences.

\noindent \textbf{Performance Analysis on Sintel Test.}
To investigate the superior performance of VideoFlow, we provide additional metrics in Table~\ref{Tab: sintel detail}, where `unmatched' refers to EPE over occluded or out-of-boundary pixels and $s_{0-10}$, $s_{10-40}$, $s_{40+}$ denote EPE over pixels with ground truth flow motion magnitude falling to $0-10$, $10-40$ and more than $40$ pixels, respectively. 
We select SKFlow*, FlowFormer, and FlowFormer++, which are the most competitive methods, for comparison.
Compared with ``matched'' pixels, ``unmatched'' pixels are hard cases because they are invisible in the target image.
Similarly, pixels whose flow motion magnitudes are larger are more challenging especially on the final pass because faster movement leads to more severe motion blur.
On the clean pass,
VideoFlow does not show performance gain over `Matched' pixels compared with FlowFormer and FlowFormer++ because these cases are rather easy.
However, VideoFlow presents dominating superiority over the other metrics that measure flows of challenging pixels: `unmatched' pixels, large-motion pixels, and even ``matched'' pixels on the final pass.
The 5-frame VideoFlow reduces 18.5\% AEPE of `Matched' pixels on the final pass from FlowFormer++.
The clear performance improvements obtained by our VideoFlow on unmatched pixels indicate that VideoFlow effectively reduces the ambiguity of out-of-view pixels with the wider integrating temporal cues.
Besides, 
our VideoFlow brings significant gains over pixels with large movements, especially on the more challenging final pass, which denotes that VideoFlow alleviates distractions from motion blurs by context information.

\subsection{Qualitative Experiment}

We visualize flow predictions of FlowFormer++~\cite{shi2023flowformer++} and our VideoFlow on Sintel and KITTI test sets in Fig.~\ref{fig:quality} to show the superior performance of VideoFlow over FlowFormer++.
By utilizing temporal cues, the blue rectangles highlight that our VideoFlow preserves more details and handles ambiguity better:
in the first row, VideoFlow shows the gaps between barriers but FlowFormer++ only produces blurry flows; in the second row, FlowFormer++ produces accident artifacts at the right of the car while VideoFlow erases them because wider temporal cues significantly improve the flow robustness. 
Moreover, FlowFormer++ fails to distinguish the shadows from the ground for the bicycle and the car while our VideoFlow predicts better flows.

\subsection{Ablation Study}

We conduct a series of ablation studies to show the effectiveness of our designs. 


\noindent \textbf{Three-frame model design.} We verify the two critical designs of our three-frame model TROF as in Table~\ref{Tab: trof design.}. We re-implement ~\cite{ren2019fusion} based on our network as baseline (the first row of Table~\ref{Tab: trof design.}), which warps the correlation features and flow predictions of the first frame pair to align with current frame pair. We first convert it to predicting bi-directional optical flows originating from the center frame (the second row of Table~\ref{Tab: trof design.}). Then we remove the independent fusion layer and fuse bi-directional motion features through the recurrent process (the third row of Table~\ref{Tab: trof design.}). Results show that the bi-directional estimation brings clear performance gains over the uni-directional baseline on most metrics. Moreover, the recurrent fusion further boosts the performance.

\noindent \textbf{Motion propagation module design.} We propose motion propagation module to bridge individual TROFs. One naive strategy is to only pass correlation features and bi-directional flow features to adjacent units (the first row in \ref{Tab: mop design.}), which has limited temporal receptive field. We propose to additionally maintain a motion state feature $\mathbf{M}_{t}^{k}$ (the third row in Table~\ref{Tab: mop design.}). In this way, the temporal receptive field grows with the recurrent updating process. We also tried adding a temporal GRU module to pass motion state feature through all TROF units in each iteration (the second row of Table~\ref{Tab: mop design.}). But this strategy brings performance drop on the FlyingThings and KITTI-2015 datasets.

\noindent \textbf{Bi-directional flows comparison.} Our VideoFlow jointly estimates bi-directional flows. In Table~\ref{Tab: bidirectional comparison.}, we compare the accuracy of bi-directional flows for both three-frame and five-frame VideoFlow models. Specifically, for the backward flow test, we pass the input image sequence in reverse order and compare the estimated backward flows with ground truth. As shown in Table~\ref{Tab: bidirectional comparison.}, the bi-directional predictions achieve similar accuracy because of the symmetry of our model. Such high-quality bi-directional flows naturally fits downstream video processing algorithms.

\setlength{\tabcolsep}{2pt}
\begin{table}[t]
\centering
\scriptsize
\begin{tabular}{cccccccc}
\hline
  {\multirow{2}{*}{Bi-directional}} &{\multirow{2}{*}{Recurrent Fusion}} 
& \multicolumn{2}{c}{Things (val)}  
 & \multicolumn{2}{c}{Sintel (train)}                    
 & \multicolumn{2}{c}{KITTI-15 (train)} \\
 \cmidrule(r{1.0ex}){3-4} \cmidrule(r{1.0ex}){5-6}
\cmidrule(r{1.0ex}){7-8}
   &  & \multicolumn{1}{c}{Clean} & \multicolumn{1}{c}{Final} & \multicolumn{1}{c}{Clean} & \multicolumn{1}{c}{Final} & \multicolumn{1}{c}{Fl-epe} & \multicolumn{1}{c}{Fl-all} \\ 
\hline
\xmark & \xmark & 2.70 & 2.53 & 1.55 & 2.62 & 4.82 & 17.48\\
\cmark & \xmark & 2.61 & 2.52 & 1.49 & 2.58 & 4.60 & 18.05\\ 
\cmark & \cmark & $\mathbf{2.54}$ & $\mathbf{2.49}$ & $\mathbf{1.48}$ & $\mathbf{2.49}$ & $\mathbf{4.51}$ & $\mathbf{16.52}$\\ 
\hline \\
\end{tabular}
\caption{\label{Tab: trof design.} $\textbf{Three-frame model design.}$ Bi-directional estimation can better utilize temporal information as motion features are well aligned in the center frame. Recurrent fusion further benefits motion features integration.
}
\end{table}

\setlength{\tabcolsep}{2pt}
\begin{table}[t]
\centering
\scriptsize
\begin{tabular}{cccccccc}
\hline
  {\multirow{2}{*}{$\mathbf{M}_{t}^{k}$?}} &{\multirow{2}{*}{Propagation Range}} 
& \multicolumn{2}{c}{Things (val)}  
 & \multicolumn{2}{c}{Sintel (train)}                    
 & \multicolumn{2}{c}{KITTI-15 (train)} \\
\cmidrule(r{1.0ex}){3-4} \cmidrule(r{1.0ex}){5-6}
\cmidrule(r{1.0ex}){7-8}
   &  & \multicolumn{1}{c}{Clean} & \multicolumn{1}{c}{Final} & \multicolumn{1}{c}{Clean} & \multicolumn{1}{c}{Final} & \multicolumn{1}{c}{Fl-epe} & \multicolumn{1}{c}{Fl-all} \\ 
\hline
\xmark & Adjacent Units & 1.61 & 1.43 & 1.15 & 2.53 & 4.02 & 14.68\\
\cmark & All Units & 1.56 & 1.40 & $\mathbf{1.07}$ & $\mathbf{2.47}$ & 4.04 & 14.55\\ 
\cmark & Adjacent Units & $\mathbf{1.48}$ & $\mathbf{1.36}$ & 1.16 & 2.56 & $\mathbf{3.89}$ & $\mathbf{14.2}$\\
\hline \\
\end{tabular}
\caption{\label{Tab: mop design.} $\textbf{Motion propagation design.}$ Our motion propagation module maintains a motion state feature $\mathbf{M}_{t}^{k}$ which absorbs adjacent units' motion state features and integrates wider temporal cues over iterations (the third row).
}
\end{table}

\setlength{\tabcolsep}{2pt}
\begin{table}[t]
\centering
\scriptsize
\begin{tabular}{cccccccc}
\hline
  {\multirow{2}{*}{Frame Number}} &{\multirow{2}{*}{Flow Direction}} 
& \multicolumn{2}{c}{Things (val)}  
 & \multicolumn{2}{c}{Sintel (train)}                    
 & \multicolumn{2}{c}{KITTI-15 (train)} \\
\cmidrule(r{1.0ex}){3-4} \cmidrule(r{1.0ex}){5-6}
\cmidrule(r{1.0ex}){7-8}
 &  & \multicolumn{1}{c}{Clean} & \multicolumn{1}{c}{Final} & \multicolumn{1}{c}{Clean} & \multicolumn{1}{c}{Final} & \multicolumn{1}{c}{Fl-epe} & \multicolumn{1}{c}{Fl-all} \\ 
\hline
3 & Forward & 1.62 & 1.42 & 1.03 & 2.19 & 3.96 & 15.33\\
3 & Backward & 1.63 & 1.42 & 0.98 & 2.23 & 4.05 & 14.74\\ 
5 & Forward & 1.48 & 1.36 & 1.16 & 2.56 & 3.89 & 14.2\\
5 & Backward & 1.49 & 1.37 & 1.20 & 2.66 & 3.79 & 14.44\\
\hline \\
\end{tabular}
\caption{\label{Tab: bidirectional comparison.} $\textbf{Forward and backward flows comparison.}$ Our VideoFlow predicts multi-frame bi-directional flows, naturally fitting downstream video processing algorithms.
}
\end{table}

\section{Conclusion}

We propose VideoFlow, which takes TRi-frame Optical Flow (TROF) module as building block in a three-frame manner. We further extend it to handle more frames by bridging TROF units with motion propagation module. Our method outperforms previous methods with large margins on all benchmarks.

\section{Acknowledgements}
This project is funded in part by National Key R\&D Program of China Project 2022ZD0161100, by the Centre for Perceptual and Interactive Intelligence (CPII) Ltd under the Innovation and Technology Commission (ITC)’s InnoHK, by General Research Fund of Hong Kong RGC Project 14204021. Hongsheng Li is a PI of CPII under the InnoHK.

{\small
\bibliographystyle{ieee_fullname}
\bibliography{egbib}
}

\end{document}